\documentclass[conference]{IEEEtran}
\IEEEoverridecommandlockouts

\usepackage[T1]{fontenc}   
\usepackage{cite}
\usepackage{amsmath,amssymb}
\usepackage{graphicx}
\usepackage{adjustbox}
\usepackage{eurosym}
\usepackage[section]{placeins}
\usepackage{dblfloatfix}
\usepackage{booktabs}
\usepackage{balance}
\usepackage[hidelinks]{hyperref}
\graphicspath{{./}}
\begin{document}

\title{An Empirical Study of Wrist-Target Construction\\ and Joint-Angle Limits in Humanoid Arm Retargeting}

\author{\IEEEauthorblockN{MinJun Chang$^{1}$ }
\thanks{$^{1}$Georgia Institute of Technology
        {\tt\small minjunchang@gatech.edu}}%
        }

\maketitle
\IEEEpeerreviewmaketitle
\begin{abstract}
Motion retargeting maps a demonstrated human movement onto a robot to compute its joint motion. This study examines how the construction of the desired wrist position affects whether computed arm joint angles reach their permitted limits on a Unitree H1-2. A reconstructed target is formed from demonstrated arm directions and robot segment lengths, so its shoulder-to-wrist distance depends on the demonstrated angle between the segments even when no posture-matching preference is applied. A reach-normalized target retains that direction while rescaling the distance, using the same processed demonstrations and the same joint-angle method. Across a forward-reach sweep the reconstructed target places a joint angle at a bound in 10 of 20 repetitions and the reach-normalized target in none. Across 25 recordings from one participant the corresponding counts are 8 and 6 on the reaching arm and 14 and 5 on the supporting arm, and a second participant repeats the bricklaying and lifting comparison. In hardware replay a joint is commanded to a bound in 12 of 16 direction-matching arm-records and in none of the 16 normalized records, and within the eligible records both tracking error and executed torque at those frames exceed their values elsewhere in the same run. Cross-body reach and arm-to-arm clearance identify limits of radial adjustment. The findings support treating target construction and posture matching as separate retargeting decisions, and evaluating both alongside the wrist positions and clearance that manipulation requires.
\end{abstract}

\begin{center}
\footnotesize\emph{This work has been submitted to the IEEE for possible publication.
Copyright may be transferred without notice, after which this version may
no longer be accessible.}
\end{center}

\section{Introduction}

Humanoid robots have attracted interest for construction and manufacturing operations in workspaces designed around human workers~\cite{liu2026construction}. Their potential to handle materials and use existing tools depends on generating motions that accommodate the robot's arm lengths and joint constraints. Human demonstrations provide observed movement sequences and reach directions for generating such motions. Motion retargeting maps a demonstrated movement onto the robot to compute its joint motion~\cite{gleicher1998}. However, differences in human and robot arm lengths and joint structure require the demonstrated movement to be adapted rather than copied directly.

Each robot joint has a permitted angular range. A retargeted motion can place a computed joint angle at the boundary of this range without exceeding it. At that configuration, the joint has no remaining permitted rotation in one direction. Remaining within the joint bounds therefore does not fully characterize the generated motion. This motivates examining which retargeting choices produce such configurations and how the robot follows the resulting commands. Tracking error and recorded joint torque provide additional information about execution.

Retargeting can specify where the wrist should move separately from how closely the arm should resemble the demonstration. The wrist target is the desired three-dimensional wrist position. Posture matching encourages the robot's shoulder-to-elbow and elbow-to-wrist directions to align with those observed in the demonstration. Removing this preference does not alter a target already constructed from these directions using robot arm lengths. As the directions approach alignment, the requested shoulder-to-wrist distance approaches the combined upper-arm and forearm length. Arm extension expresses this distance relative to that total length.

This study examines how wrist-target construction affects whether computed arm joint angles reach their permitted limits on a Unitree H1-2. A reconstructed target uses demonstrated arm directions and robot arm lengths. A reach-normalized target retains the resulting shoulder-to-wrist direction while rescaling its distance. Both use the same processed demonstrations and method for computing joint angles without posture matching. This isolates target construction within the same arm model. Evaluation covers five motions, a forward-reach sweep, a second participant, additional posture and forearm-direction comparisons, and hardware replay of prepared trajectories.

The contribution is empirical evidence that wrist-target construction can affect whether computed joint angles reach their limits even when posture matching is absent. The findings support treating target construction and posture matching as separate retargeting decisions. They also distinguish fewer computed angles at their limits from preservation of the wrist positions and arm clearance required for manipulation.

\section{Related Work}

Generating robot motion from a human demonstration involves calculating the desired motion and controlling the robot to follow it. H2O uses body-point positions scaled to the robot~\cite{loper2015smpl} in whole-body teleoperation~\cite{he2024h2o}. HumanPlus combines shadowing of human motion with subsequent skill learning~\cite{fu2024humanplus}. OmniH2O extends teleoperation and learning to dexterous whole-body behaviors~\cite{he2024omnih2o}. BeyondMimic learns a tracking policy from a corpus of human motion capture and composes the tracked skills into new behaviors~\cite{liao2025beyondmimic}. ASAP reduces the mismatch between simulated and physical dynamics so that a tracking policy transfers to hardware~\cite{he2025asap}. These systems treat the robot's joint bounds as constraints on the executed motion. The question examined here is which retargeting choices bring computed angles to those bounds, not whether the bounds are respected.

Other approaches adjust demonstrated motion using the robot's geometry or the surrounding environment. OKAMI uses object locations to modify trajectories before computing joint angles~\cite{li2024okami}. OmniRetarget preserves spatial and contact relationships with the environment~\cite{yang2025omniretarget}. ReActor combines physics-aware motion adaptation with the training of a tracking policy~\cite{muller2026reactor}. GMR is the closest in subject, since it addresses errors introduced when a source motion is scaled to a robot and evaluates the effect on downstream tracking~\cite{araujo2025gmr}. The present study differs in scope rather than in aim. It holds one joint-angle method fixed, varies only the requested shoulder-to-wrist distance, and reports counts of computed angles at their bounds on one arm-only platform. A prior construction-focused workshop study compares human-derived and robot-specific motions between matched targets, and supplies the direction-matching objective used here as an additional comparison~\cite{ours_iros2026ws}.

An arm may allow different joint configurations to reach the same wrist position. Redundancy resolution uses this freedom to adjust posture while retaining the requested position locally~\cite{whitney1969,liegeois1977}. Joint-limit avoidance uses these adjustments to favor angles away from their permitted bounds~\cite{chan1995}. Damped least squares supports joint-angle computation near configurations where wrist movement becomes difficult in some directions~\cite{nakamura1986,wampler1986}. This study keeps the joint-angle method fixed while changing the desired wrist position. A separate comparison examines joint centering that competes with wrist-position accuracy rather than using position-preserving adjustments.

\section{Problem Formulation}

\subsection{Arm joints and available motion}

The comparison uses four active joints per arm on the H1-2. The three shoulder joints control pitch, roll and yaw. Elbow flexion provides the fourth joint. All three wrist joints remain fixed at zero. This setup compares requested wrist positions within the same arm model rather than evaluating full hand poses or whole-body movement. Human arm directions are expressed relative to a torso coordinate frame defined by shoulder and hip keypoints. This relative representation does not require camera-to-robot extrinsic calibration.

The wrist-position Jacobian maps small joint-angle changes to wrist-position changes. Across both arms of all 25 recordings it has numerical rank three on $100.00\%$ of frames. With rank three and four active joints, it leaves one local direction of joint motion that preserves wrist position to first order~\cite{tolani2000}. This is the available redundancy before additional constraints. Near-full extension can worsen numerical conditioning. Joint bounds further restrict the usable adjustments, motivating comparison of the requested wrist distances.

\subsection{Extension retained in the desired wrist position}

The upper-arm and forearm unit vectors point from shoulder to elbow and elbow to wrist. Each is multiplied by the corresponding robot segment length. Their sum gives the reconstructed shoulder-to-wrist vector in \eqref{eq:recon}. Its length depends on the demonstrated angle between the segments \eqref{eq:ext}. Removing posture matching does not change this calculation. Reconstructed extension divides the vector's length by total robot arm length rather than dividing human reach by human arm length.
\begin{equation}
v_t = \ell_u \hat{u}_t + \ell_f \hat{f}_t, \quad
x_t^{\text{rec}} = a_t + v_t, \quad
e_t = \frac{\|v_t\|}{L}, \quad
\hat{d}_t = \frac{v_t}{\|v_t\|}
\label{eq:recon}
\end{equation}
where $a_t$ is the shoulder position and $L$ is total robot arm length. The vector $v_t$ gives the requested shoulder-to-wrist direction and distance, and $\hat{d}_t$ is the retained direction.
\begin{equation}
e_t^2 = \frac{\ell_u^2 + \ell_f^2 + 2\,\ell_u \ell_f\, \hat{u}_t^{\!\top} \hat{f}_t}{L^2}
\label{eq:ext}
\end{equation}

\subsection{Two ways to calculate the wrist position}

\begin{figure}[!t]
\centering
\includegraphics[width=\columnwidth,height=0.30\textheight,keepaspectratio]{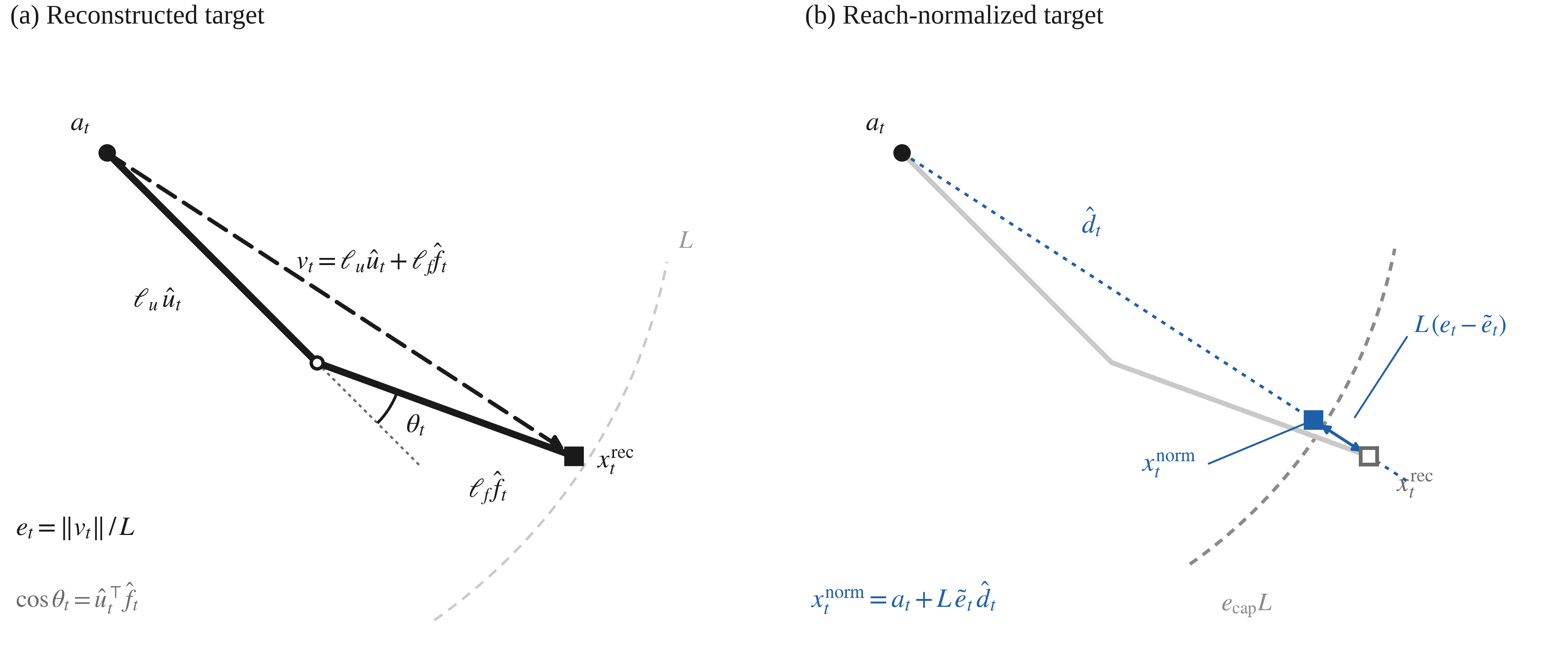}
\caption{Reconstructed and reach-normalized wrist targets share a shoulder-to-wrist direction but differ in distance. The reconstructed distance depends on the demonstrated angle between the arm segments in \eqref{eq:recon} and \eqref{eq:ext}. Normalization rescales that distance along the same direction and applies the selected cap.}
\label{fig:targets}
\end{figure}

Reach normalization retains the reconstructed reach direction while rescaling the shoulder-to-wrist distance (Fig.~\ref{fig:targets}). One scale per arm and recording maps the 95th-percentile extension to a cap of $0.88$. Larger scaled values are clipped. A cap of $0.84$ tests sensitivity. Relative extension is preserved only where clipping is inactive. Scaling may increase some reaches and decrease others, so it is not uniform shortening. It changes wrist positions and the distance between the hands.
\begin{equation}
s = \frac{e_{\text{cap}}}{P_{95}(\{e_t\})}, \quad
\tilde{e}_t = \min(s\,e_t,\; e_{\text{cap}}), \quad
x_t^{\text{norm}} = a_t + L\,\tilde{e}_t\, \hat{d}_t
\label{eq:norm}
\end{equation}
A scale of $s = 1$ is used when the percentile falls below $10^{-6}$.

\subsection{Primary and additional comparisons}

The primary comparison asks whether changing the desired wrist distance affects how often computed joint angles reach their permitted limits. Both targets use the same joint-angle method and processed input, and their secondary centering and continuity preferences remain fixed. The additional comparisons in Sections~\ref{sec:dirpost} and~\ref{sec:forearm} instead change the criteria for computing joint angles. Hardware replay changes both choices.

\section{Methods}

\subsection{Motion capture and preprocessing}

Motion is recorded with a neck-mounted ZED stereo camera while the robot remains stationary~\cite{stereolabs_zed}. The body tracker estimates a skeleton with 38 keypoints from 720p images at 60 frames per second. Each recording is trimmed to the motion interval. Keypoints with rescaled confidence below $0.3$ are excluded. A direction-change gate rejects changes above 600 degrees per second and relocks after five consecutive rejections, and a One-Euro filter smooths the directions with a minimum cutoff of $1.0$ and a speed coefficient of $0.5$~\cite{casiez2012oneeuro}.

Far-reaching recordings contained estimated shoulder-to-wrist distances of 72 to 79 cm for an arm measuring approximately 55.9 cm. Each skeleton is therefore normalized to the participant's measured length of either $0.5588$ m or $0.4953$ m. Processing retains the shoulder position and reach direction. It limits the reach distance to the measured arm length and recalculates the elbow position using a two-link arm model. The elbow remains on the side indicated by the measured pose. Every retargeting condition receives the same processed input.

\subsection{Joint-angle calculation for both wrist targets}

\begin{table}[!t]
\centering
\caption{Limits of the four active joints per arm. Effort and velocity are model parameters, not constraints on the offline solve.}
\label{tab:joints}
\small
\setlength{\tabcolsep}{5pt}
\renewcommand{\arraystretch}{1.08}
\begin{tabular}{lcccc}
\toprule
Joint & Right range & Left range & Effort & Velocity \\
      & (rad) & (rad) & (N\,m) & (rad/s) \\
\midrule
Shoulder pitch & $[-3.14,\ 1.57]$ & $[-3.14,\ 1.57]$ & 40 & \phantom{0}9 \\
Shoulder roll  & $[-3.40,\ 0.38]$ & $[-0.38,\ 3.40]$ & 40 & \phantom{0}9 \\
Shoulder yaw   & $[-3.01,\ 2.66]$ & $[-2.66,\ 3.01]$ & 18 & 20 \\
Elbow          & $[-0.95,\ 3.18]$ & $[-0.95,\ 3.18]$ & 18 & 20 \\
\bottomrule
\end{tabular}
\end{table}

Both wrist targets use the same inverse-kinematics method to compute robot joint angles. A damped least-squares solver reduces the difference between the desired wrist position and the position calculated from the current angles~\cite{nakamura1986,wampler1986}. This difference is the wrist-position residual. Secondary preferences favor joint angles near the middle of their ranges and continuity with the preceding solution. A null-space projector selects adjustments that preserve wrist position in the local linear approximation. Centering uses squared distance from each joint midpoint normalized by its half-range. Continuity penalizes changes from the preceding accepted solution. No manipulability term is active.
\begin{align}
N &= I - J^{+} J \label{eq:sec} \\
U(q) &= \textstyle\sum_j (q_j - m_j)^2 / r_j^2 \nonumber \\
[\nabla U(q)]_j &= 2 (q_j - m_j) / r_j^2 \nonumber \\
\Delta q_{\text{sec}} &= -N\left[\lambda_c\,\nabla U(q) + \lambda_p\,(q - q_{\text{prev}})\right] \nonumber
\end{align}
Here $q$ is the current joint vector and $q_{\text{prev}}$ the preceding accepted solution. Each $m_j$ is a joint midpoint and $r_j$ is its half-range plus $10^{-6}$. The matrix $J$ is the active wrist-position Jacobian and $J^{+}$ its pseudoinverse. The projected centering and continuity coefficients are fixed at $\lambda_c = 0.05$ and $\lambda_p = 0.02$ in every condition of the primary comparison.

The solver uses a damping parameter of $10^{-2}$, a wrist-position tolerance of $10^{-4}$ m and at most 100 iterations per frame. After each update, joint angles are clipped to their permitted ranges. Finite updates and clipping can still affect wrist-position error despite the local projection. Neither the continuity preference nor the replay speed scale imposes a joint-velocity limit in the offline solve, and the solver does not enforce torque limits.

Each recording starts from a fixed rest pose, and later frames use the preceding accepted solution without random restarts. In the joint order of Table~\ref{tab:joints}, the initial angles are $(0.30, -0.24, 0.12, 0.69)$ rad on the right and $(0.31, 0.24, -0.12, 0.67)$ rad on the left. Both arms start with a minimum half-range-normalized distance to a bound of approximately $0.328$. The effort and velocity columns are model parameters not enforced by this calculation.

\subsection{Additional comparisons: direction and posture matching}
\label{sec:dirpost}

Direction matching computes joint angles by aligning the robot's upper-arm and forearm directions with the demonstrated directions, and is the objective of our prior workshop study~\cite{ours_iros2026ws}. Damped Gauss-Newton optimization minimizes the equally weighted differences between these unit vectors. Regularized direction matching adds shoulder-yaw preferences with yaw-regularization and yaw-neutral coefficients of $0.05$ and $0.02$. Shoulder yaw is held fixed when elbow flexion is below 12 degrees. Neither condition calculates a separate desired wrist position, so both differ from the primary comparison in how motion is specified and how joint angles are selected.

The H2O-inspired condition matches robot-scaled elbow and wrist positions with equal weight. The OKAMI-inspired condition combines wrist-position error with upper-arm and forearm direction errors. Their multipliers are $0.04g$ and $0.08g$ before squaring. Position errors are in meters and direction errors are dimensionless. A separate two-by-four study combines posture gains of $1$ and $0.001$ with competing centering weights $w_c$ of $0$, $0.05$, $0.20$ and $0.50$. The weight $w_c$ enters the residual vector alongside wrist-position error, so centering competes with position accuracy rather than acting through the null-space projection of \eqref{eq:sec}. It is therefore a different quantity from $\lambda_c$, which remains at $0.05$ throughout the primary comparison, and the shared numerical value at $w_c = 0.05$ is a coincidence rather than a correspondence. The lower posture gain is not zero. It gives direction multipliers of $4 \times 10^{-5}$ and $8 \times 10^{-5}$. These conditions adapt selected objectives rather than reproduce the complete H2O and OKAMI systems.
\begin{equation}
r_g(q) = \begin{bmatrix} p_w(q) - x_t^{\text{rec}} \\ 0.04g\,(u_r(q) - \hat{u}_t) \\ 0.08g\,(f_r(q) - \hat{f}_t) \end{bmatrix}
\label{eq:posture}
\end{equation}
The robot wrist position is $p_w(q)$, the directions $u_r(q)$ and $f_r(q)$ are the robot upper arm and forearm, and the gain $g$ scales the posture errors.

\subsection{Additional forearm-direction requirement}
\label{sec:forearm}

An additional test examines the effect of requiring the robot forearm to follow the demonstrated direction while reaching the reconstructed wrist target. Forearm-direction error is added with a multiplier of $0.5$. Its Jacobian is estimated by perturbing joint angles by $10^{-4}$ rad. The requirement is omitted, applied at every frame or enabled around low-speed intervals. The gate selects local wrist-speed minima below the recording's 25th percentile. Selected minima are separated by at least $0.30$ s. Each opens a $0.10$ s window on either side. These intervals are selected from speed rather than independently observed contact events.
\begin{equation}
r_a = \begin{bmatrix} x_t^{\text{rec}} - p_w(q) \\ 0.5\,(\hat{f}_t - f_r(q)) \end{bmatrix}, \qquad
J_a = \begin{bmatrix} J & 0.5\,J_f \end{bmatrix}
\label{eq:gate}
\end{equation}
The combined Jacobian $J_a$ defines the projector while the gate is active. If it has column rank four, no local joint motion preserves both position and direction.

\section{Experimental Design}

\subsection{Demonstrations and dataset assignment}

Two participants provided laboratory demonstrations. The first performed five repetitions each of overhead hammering, bricklaying, lifting, two-handed pick-and-stack and cross-body reach. These 25 recordings span overhead reaching, material handling and cross-body reaching. The second participant performed five bricklaying and five lifting recordings to repeat part of the comparison. Both arms are evaluated separately. Arms from one recording and repeated replay speeds are not treated as independent human demonstrations.

The primary wrist-target comparison uses all 25 recordings from the first participant and all 10 from the second. The posture-weight, elbow-and-wrist position and forearm-direction tests use the first participant's 25 recordings. The forward-reach sweep uses four chest-height distances of 12, 16, 20 and 24 inches at five repetitions each, and only the first participant's sweep provides a usable extension contrast. Hardware replay uses the four recordings described in Section~\ref{sec:replay}. These datasets support comparisons within the evaluated movements rather than generalization across workers.

\subsection{Saturation and wrist-position error}

Computed joint angles are evaluated after clipping to their permitted ranges. An offline frame is termed \emph{saturated} when any active joint angle lies within $10^{-6}$ rad of a bound, and a recording is counted as saturated if at least one such frame occurs. No minimum duration is required. A hardware frame is termed saturated when a commanded joint angle has a normalized margin at or below $10^{-4}$, where that margin is zero at a joint bound and one at mid-range. The two thresholds have different units and are not interchangeable. Saturation as defined here denotes an angle at a bound after clipping. It identifies neither a command outside the permitted range nor physical contact with a mechanical stop.

Wrist-target error is the difference between the desired wrist position and the position calculated from the solved joint angles. It measures numerical accuracy or the compromise between competing position and posture preferences. Because the two primary conditions have different desired positions, their errors do not measure agreement with the same wrist reference. Section~\ref{sec:weights} compares errors while retaining the same target and changing objective weights. Forearm-direction error is evaluated separately for the speed-gated directional test.

\subsection{Hardware replay protocol}
\label{sec:replay}

Hardware replay examines how the robot follows commands when a joint is requested to reach its angular limit. It compares regularized direction matching with reach normalization. These conditions differ in both the requested motion and the joint-angle calculation. The hardware comparison therefore does not isolate the effect of changing the wrist target. One bricklaying and three lifting recordings are retargeted under both conditions and replayed at speed scales of $0.2$ and $0.5$, producing 16 bimanual runs and 32 arm-records. An arm-record is one arm in one run, so each condition contributes 16.

A replay set is retained when both arms are recorded and no commanded joint exceeds its permitted range. Earlier single-arm sets fall outside this protocol. One contains out-of-range commands and the other lacks the field needed to verify compliance. Both retained conditions use the same gravity feed-forward settings. A 2 Hz low-pass filter is applied after inverse kinematics and followed by joint-range clipping. Bar attachments are removed and torso-clearance corrections are active. The hardware results therefore describe these prepared trajectories rather than the unmodified references.

For each arm-record, tracking error is the mean absolute difference between commanded and measured angles across four active joints. The torque quantity is mean absolute executed torque across the same joints. Within-run ratios compare the mean over saturated frames against the mean over the remaining frames. Both frame sets must be nonempty and the denominator must be nonzero. Torque analysis requires at least 20 frames in each set. The reported correlations are descriptive Pearson coefficients.
\begin{align}
E_t &= \tfrac{1}{4}\textstyle\sum_{j=1}^{4} |q^{\text{exe}}_{tj} - q^{\text{cmd}}_{tj}|, \qquad
T_t = \tfrac{1}{4}\textstyle\sum_{j=1}^{4} |\tau^{\text{exe}}_{tj}| \label{eq:ratios} \\
R_E &= \frac{\operatorname{mean}_{t \in S}(E_t)}{\operatorname{mean}_{t \notin S}(E_t)}, \qquad
R_T = \frac{\operatorname{mean}_{t \in S}(T_t)}{\operatorname{mean}_{t \notin S}(T_t)} \nonumber
\end{align}
The set $S$ contains the saturated frames of the record.

\subsection{Base-position and arm-clearance checks}

Both arms must use the same robot base position. The offline calculation therefore selects one horizontal translation per recording that minimizes the largest radial deficit across both arms. A radial deficit is the distance by which a demonstration-derived wrist reference exceeds a radius of $0.88L$ from the translated shoulder. Results are compared with an exploratory 8 cm threshold. A separate check evaluates arm-to-arm clearance with $0.20$ m bar attachments against an 8 cm criterion, and torso intersection is also examined. The two 8 cm values concern different properties and neither is a manipulation-success measure.

\section{Results}

\subsection{Primary comparison: desired wrist positions}

\begin{table*}[!t]
\centering
\caption{Saturated recordings, out of five repetitions per motion. Reaching arm, supporting arm in parentheses.}
\label{tab:primary}
\small
\setlength{\tabcolsep}{5pt}
\renewcommand{\arraystretch}{1.08}
\begin{adjustbox}{max width=\textwidth}
\begin{tabular}{lcccc}
\toprule
Motion & Direction matching & Regularized & Reconstructed & Reach-normalized \\
\midrule
\multicolumn{5}{l}{\emph{First participant: five repetitions per motion}} \\
Overhead hammering & 0/5 (0/5) & 0/5 (0/5) & 0/5 (0/5) & 0/5 (0/5) \\
Bricklaying        & 5/5 (5/5) & 5/5 (5/5) & 2/5 (5/5) & 0/5 (1/5) \\
Lifting            & 5/5 (3/5) & 5/5 (4/5) & 0/5 (4/5) & 0/5 (0/5) \\
Pick-and-stack     & 5/5 (5/5) & 2/5 (5/5) & 1/5 (5/5) & 1/5 (4/5) \\
Cross-body reach   & 5/5 (0/5) & 5/5 (0/5) & 5/5 (0/5) & 5/5 (0/5) \\
Total              & 20/25 (13/25) & 17/25 (14/25) & 8/25 (14/25) & 6/25 (5/25) \\
\midrule
\multicolumn{5}{l}{\emph{Second participant: five repetitions per motion}} \\
Bricklaying        & 5/5 (5/5) & 4/5 (5/5) & 1/5 (4/5) & 0/5 (1/5) \\
Lifting            & 5/5 (1/5) & 4/5 (0/5) & 1/5 (1/5) & 0/5 (0/5) \\
Total              & 10/10 (6/10) & 8/10 (5/10) & 2/10 (5/10) & 0/10 (1/10) \\
\bottomrule
\end{tabular}
\end{adjustbox}
\end{table*}

Across the first participant's 25 recordings, reach normalization reduces saturated recordings from 8 to 6 on the reaching arm and from 14 to 5 on the supporting arm, relative to the reconstructed target (Table~\ref{tab:primary}). Both conditions use the same joint-angle method and processed input, and the larger reduction occurs on the supporting arm. Direction matching produces 20 saturated reaching-arm recordings. Its difference from reach normalization combines a change in the desired motion with a change in the joint-angle calculation.

For the first participant's bricklaying recordings, the reconstructed-to-normalized reaching-arm count falls from 2/5 to 0/5. The supporting-arm count falls from 5/5 to 1/5. Lifting remains at 0/5 on the reaching arm while the supporting-arm count changes from 4/5 to 0/5. No saturation occurs during overhead hammering under any condition. Across the second participant's bricklaying and lifting recordings, counts fall from 2/10 to 0/10 on the reaching arm and from 5/10 to 1/10 on the supporting arm. The corresponding first-participant supporting-arm change is 9/10 to 1/10.

Adjusting the requested wrist distance does not resolve every motion. Cross-body reach saturates in all five reaching-arm recordings under every condition. Reach-normalized pick-and-stack retains saturated frames in one reaching-arm recording and four supporting-arm recordings. An independently implemented damped least-squares solver produces the same reconstructed-target counts of 8/25 and 14/25, which concerns the recorded counts rather than the full joint trajectories.

\subsection{Forward-reach sweep}

\begin{figure}[!t]
\centering
\includegraphics[width=\columnwidth,height=0.30\textheight,keepaspectratio]{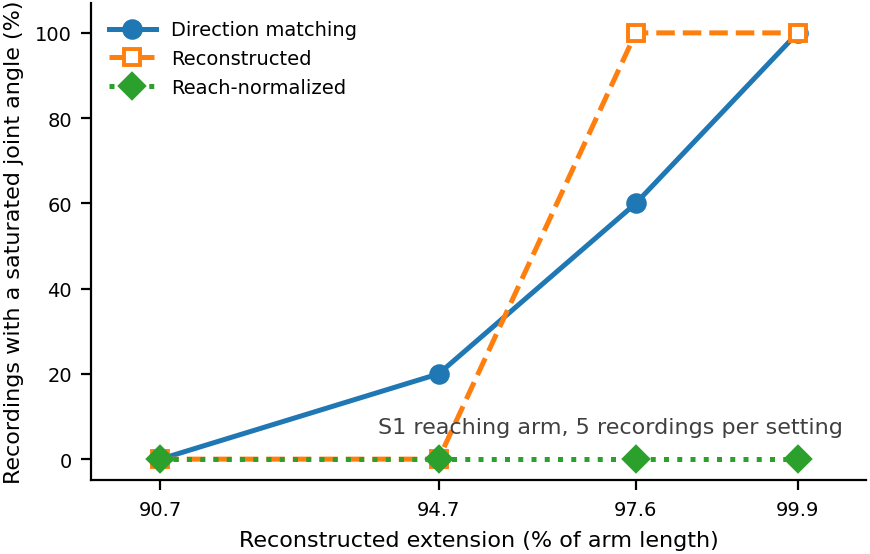}
\caption{Saturated reaching-arm recordings across four forward-reach settings. Each setting has five repetitions. Reconstructed and reach-normalized targets use the same joint-angle method. Direction matching is an additional comparison. Lines connect the evaluated settings rather than identify a precise onset threshold.}
\label{fig:sweep}
\end{figure}

Reconstructed extension rises from $90.7\%$ to $99.9\%$ across the four sweep settings (Fig.~\ref{fig:sweep}). Saturated reaching-arm recordings number 0/5, 0/5, 5/5 and 5/5 under the reconstructed target. None occurs under reach normalization. Direction matching gives 0/5, 1/5, 3/5 and 5/5. The observed change for the reconstructed target lies between the sampled $94.7\%$ and $97.6\%$ settings. These samples do not identify a precise onset threshold.

Numerical conditioning also varies with extension. The smallest-singular-value median of the wrist-position Jacobian is $0.164$ below $0.80$ extension and $0.016$ above $0.97$, which is consistent with poorer conditioning near full extension in the evaluated arm model. The normalized sweep follows its own desired wrist position to approximately $0.1$ mm, an accuracy against the modified target rather than preservation of the original wrist path. Repeating the primary comparison with a cap of $0.84$ gives the same recording counts as $0.88$. This agreement applies to the counts at these two settings rather than establishing that the cap has no effect.

\subsection{Additional comparisons: posture weights and forearm direction}
\label{sec:weights}

\begin{table}[!t]
\centering
\caption{Saturated recordings under competing posture and centering weights. Reaching arm, supporting arm in parentheses.}
\label{tab:weights}
\small
\setlength{\tabcolsep}{5pt}
\renewcommand{\arraystretch}{1.08}
\begin{tabular}{lcc}
\toprule
Competing centering weight $w_c$ & $g = 1$ & $g = 0.001$ \\
\midrule
$0$    & 23/25 (15/25) & 10/25 (15/25) \\
$0.05$ & 16/25 (15/25) & \phantom{0}7/25 (12/25) \\
$0.20$ & \phantom{0}6/25 (\phantom{0}9/25) & \phantom{0}5/25 (\phantom{0}8/25) \\
$0.50$ & \phantom{0}0/25 (\phantom{0}0/25) & \phantom{0}0/25 (\phantom{0}0/25) \\
\bottomrule
\end{tabular}
\end{table}

\begin{table}[!t]
\centering
\caption{Wrist-target error in millimeters at the two largest competing centering weights.}
\label{tab:wristerr}
\small
\setlength{\tabcolsep}{5pt}
\renewcommand{\arraystretch}{1.08}
\begin{tabular}{lccc}
\toprule
Arm & Posture gain $g$ & $w_c = 0.20$ & $w_c = 0.50$ \\
\midrule
Right & $1$     & 40.37 & 149.88 \\
Right & $0.001$ & 39.00 & 153.38 \\
Left  & $1$     & 35.03 & 173.33 \\
Left  & $0.001$ & 36.04 & 177.93 \\
\bottomrule
\end{tabular}
\end{table}

Changing the objective weights also changes how often saturation occurs. With no competing centering residual, reducing posture gain from $1$ to $0.001$ lowers reaching-arm counts from 23/25 to 10/25 (Table~\ref{tab:weights}). The supporting-arm count remains 15/25. At $w_c = 0.05$ the reaching-arm counts are 16/25 and 7/25. Larger values of $w_c$ reduce these counts at both gains. The posture-gain sweep gives 23/25, 15/25, 10/25 and 10/25 at gains of $1$, $0.1$, $0.01$ and $0.001$. The adapted elbow-and-wrist position objective gives 20/25.

The two largest values of $w_c$ produce fewer saturated recordings but greater wrist-position errors (Table~\ref{tab:wristerr}). Errors range from $35.0$ to $40.4$ mm at $w_c = 0.20$ and from $149.9$ to $177.9$ mm at $w_c = 0.50$. The computed wrist therefore departs further from its unchanged target. The low-gain weighted condition also gives 7/25 rather than the reconstructed condition's 8/25. These are different formulations. The former retains a small posture preference and lets centering compete with position accuracy. The latter applies centering through the position-task null space at the fixed coefficient $\lambda_c$.

Adding forearm-direction matching increases the number of saturated reaching-arm recordings. Counts change from 8/25 without this requirement to 19/25 with speed gating and 23/25 with continuous activation. The corresponding supporting-arm counts are 14/25, 13/25 and 13/25. The gate covers $35.9\%$ of frames. Forearm-direction error during those intervals is $1.8$ degrees. The additional requirement can restrict the position-preserving adjustments described in Section III-A. The rank condition for removing that freedom is stated in \eqref{eq:gate}.

\subsection{Hardware replay}

\begin{figure*}[!t]
\centering
\includegraphics[width=\textwidth]{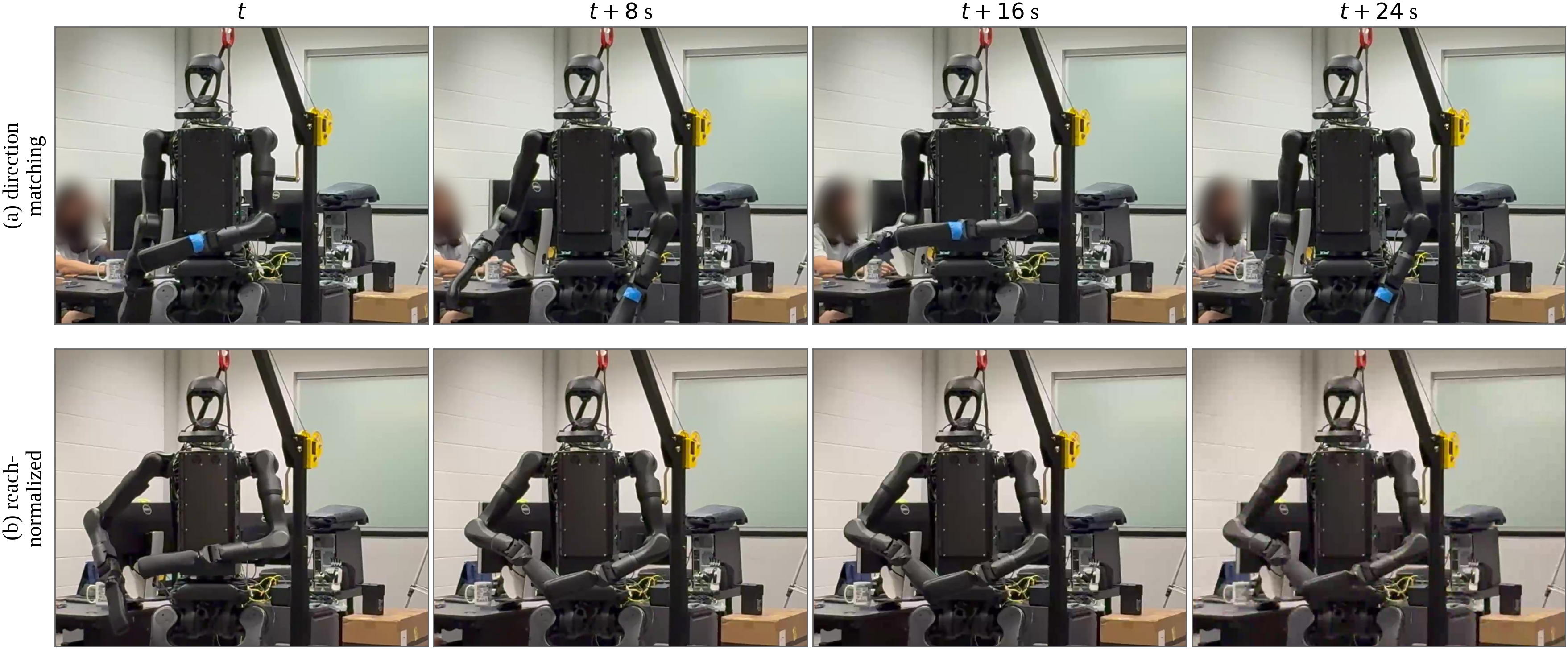}
\caption{Replayed arm motion sampled at equal intervals under the two hardware conditions of Section~\ref{sec:replay}. (a) Regularized direction matching, in which the right arm extends away from the torso and approaches full extension. (b) The reach-normalized target, in which both elbows remain flexed and the wrists stay nearer the torso. Frames are spaced $8$ s apart across $24$ s of replay in both rows, with $t$ set independently for each sequence. The two rows come from separate replay sessions and are not aligned to a common motion phase, and they differ in the requested motion and in the joint-angle method rather than in the wrist target alone. The operator's face is blurred.}
\label{fig:configs}
\end{figure*}

\begin{table}[!t]
\centering
\caption{Within-run tracking and torque associations across the 12 eligible direction-matching arm-records.}
\label{tab:hw}
\small
\setlength{\tabcolsep}{5pt}
\renewcommand{\arraystretch}{1.08}
\begin{tabular}{lcc}
\toprule
Statistic & Tracking & Torque \\
\midrule
Maximum ratio & 1.891 & 2.067 \\
Correlation with saturated fraction & 0.850 & 0.845 \\
Correlation after the paired-record omission & 0.457 & 0.602 \\
\bottomrule
\end{tabular}
\end{table}

Saturated frames occur in 12 of 16 regularized direction-matching arm-records and in none of the 16 normalized arm-records. This comparison changes both the requested motion and the joint-angle method. Fig.~\ref{fig:configs} shows the two replayed conditions at equal sampling intervals. The four direction-matching records without saturated frames come from one lifting demonstration across both arms and speeds. These four records and the 16 normalized records have no defined within-run ratio because the required frame set is empty. All 20 remain in the descriptive counts.

The same 12 eligible records enter the tracking and torque analyses, each with at least 20 frames in both torque partitions (Table~\ref{tab:hw}). The median within-run torque ratio is $1.314$ and the maximum is $2.067$, and nine ratios exceed one. The three below one occur where $1.03\%$, $1.99\%$ and $2.25\%$ of frames are saturated. Correlations between that fraction and the tracking and torque ratios are $0.850$ and $0.845$. Omitting both speeds of one bricklaying recording's left arm reduces them to $0.457$ and $0.602$ across 10 records. The omitted pair has the largest tracking ratios of $1.891$ and $1.795$. The associations therefore depend strongly on one recorded arm motion.

\begin{figure}[!t]
\centering
\includegraphics[width=\columnwidth,height=0.32\textheight,keepaspectratio]{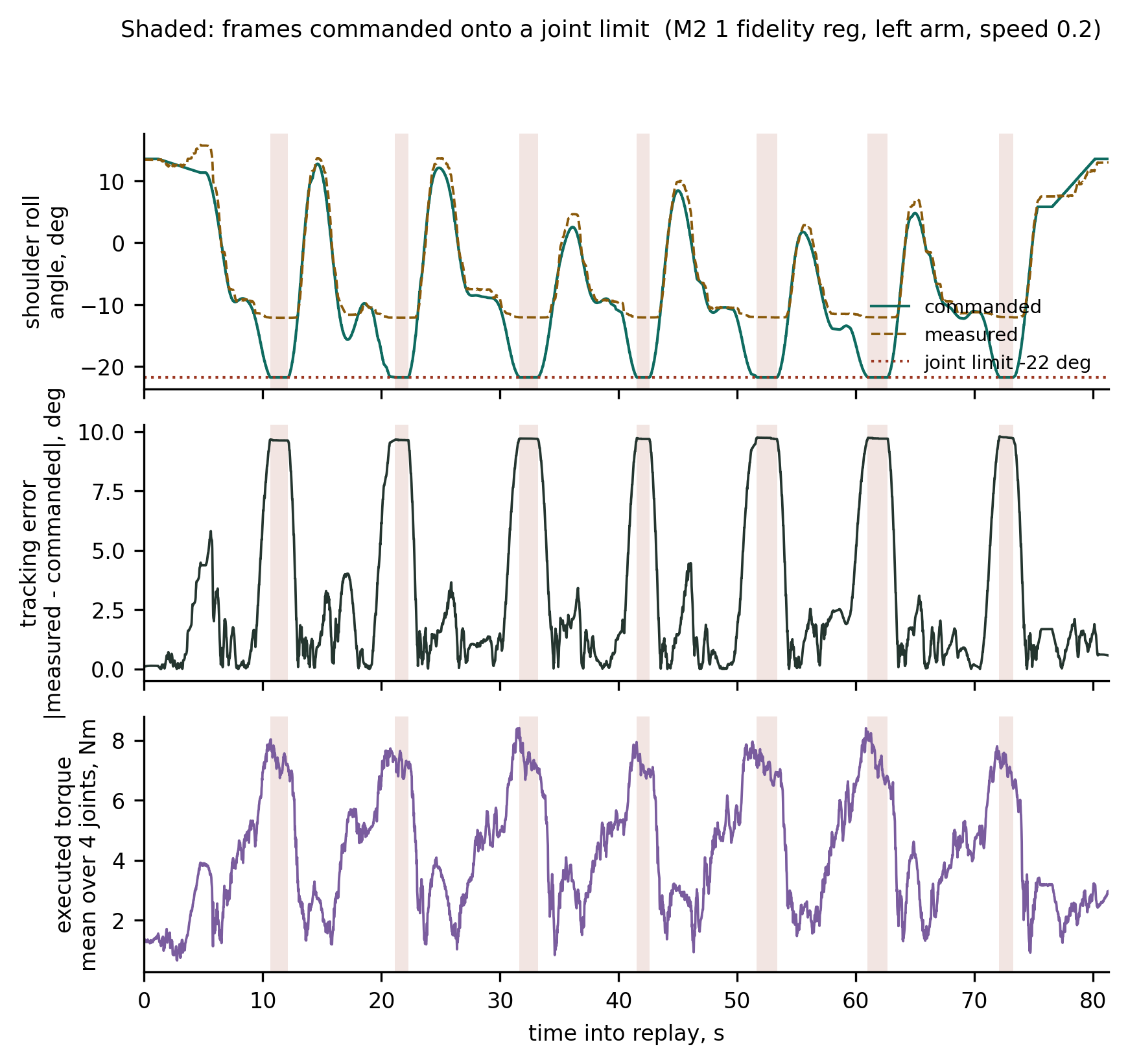}
\caption{Left-arm bricklaying replay under regularized direction matching at speed scale $0.2$. Shading marks 497 saturated frames of shoulder roll in seven episodes. The measured angle remains within the model bound. The bottom panel averages absolute executed torque across four arm joints. This recording supplies the influential pair in Table~\ref{tab:hw}.}
\label{fig:trace}
\end{figure}

Figure~\ref{fig:trace} shows the influential bricklaying replay. Shoulder roll is commanded to its $-0.380$ rad lower bound for 497 of 4068 frames in seven episodes, while the measured angle remains near $-12$ degrees. Mean absolute shoulder-roll error is $9.696$ degrees during those frames and $2.366$ degrees elsewhere, and four-joint mean absolute executed torque is $7.21$ N\,m versus $3.93$ N\,m. The arm-record ratio of $1.891$ is lower because it averages across four joints and uses 568 frames in which any joint is saturated. These values show recurring tracking discrepancies at saturated frames rather than measured contact with a mechanical stop.

\subsection{Base-position and arm-clearance results}

\begin{figure}[!t]
\centering
\includegraphics[width=\columnwidth,height=0.30\textheight,keepaspectratio]{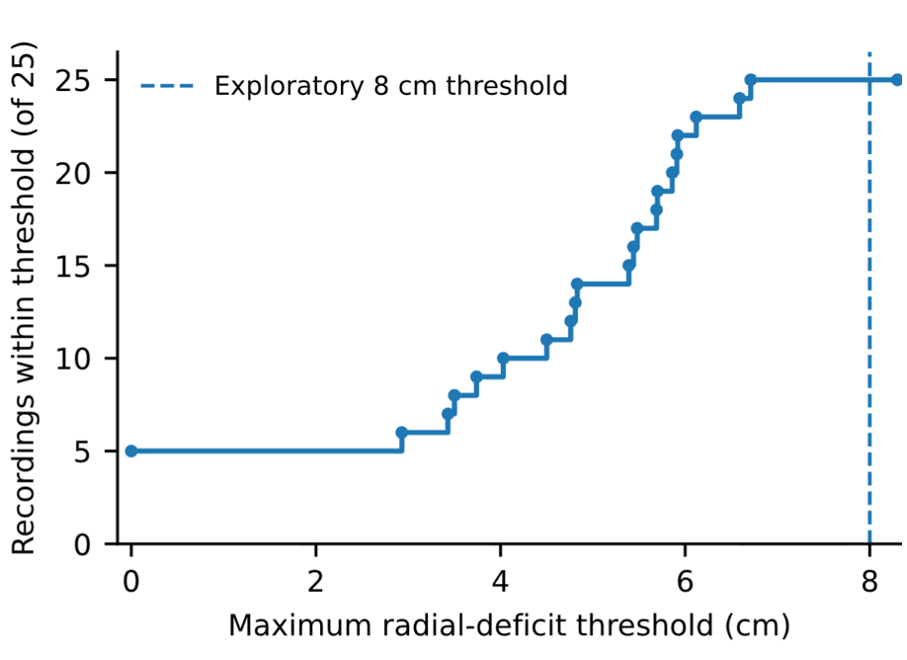}
\caption{Maximum radial deficit across both arms after one common horizontal base translation per recording. Deficits are measured relative to the capped reach radius. The 8 cm reference is an exploratory threshold rather than a validated grasp or placement tolerance.}
\label{fig:deficit}
\end{figure}

One horizontal base translation per recording keeps both arms below the exploratory 8 cm radial-deficit threshold in all 25 recordings (Fig.~\ref{fig:deficit}). The mean translation is $8.4$ cm. The mean maximum deficit is $4.05$ cm and the largest is $6.71$ cm. Both arms therefore use one base position in this calculation. The result concerns radial distances to demonstration-derived wrist references. No base translation is executed in the hardware comparison.

The arm-clearance result differs from the saturation result. With $0.20$ m bar attachments, none of the 10 normalized trajectories passes the 8 cm arm-to-arm criterion. Direction matching passes in 5 of 10. Separately, two direction-matching references with many saturated frames intersect the torso. They require wrist displacements of 32 cm and 27 cm to clear it. Fewer saturated recordings therefore do not establish feasible bimanual execution.

\section{Discussion and Limitations}

The primary comparison supports treating the calculation of desired wrist positions separately from posture matching. Reconstructed wrist distances depend on demonstrated arm directions even when no posture preference is applied during joint-angle computation. Normalizing those distances reduces saturated recordings for some evaluated motions, with the larger reduction on the supporting arm. Cross-body reach remains unchanged. These findings concern particular arms and movements on the H1-2 rather than a uniform improvement in humanoid retargeting.

The additional comparisons show that the criteria for selecting joint angles also affect the results. Reweighting posture and centering produces fewer saturated recordings but permits wrist-position errors up to $178$ mm. Adding a forearm-direction requirement increases the counts. The hardware difference of 12/16 versus 0/16 changes both the desired motion and the joint-angle method. It cannot isolate the effect of the requested wrist distance. Within-run tracking and torque observations describe execution at saturated frames. These associations remain strongly influenced by one recorded arm motion.

Changing the requested wrist distance changes hand positions and their separation. Fewer saturated recordings therefore do not demonstrate that a grasp or placement position is preserved. Using one base position for both arms reduces radial deficits without establishing the required contact geometry. The attachment-clearance result is also less favorable under normalization. Wrist-position requirements and collision clearance must consequently be evaluated alongside the saturation results before claiming successful manipulation.

The complete protocol uses one participant and one robot, and the second participant repeats only bricklaying and lifting, so results support descriptive comparison rather than population-level inference. The fixed-wrist arm model does not evaluate whole-body coordination or locomotion. Normalization requires a complete recording and is restricted to offline motion generation. Input depends on stereo pose estimation and participant-specific processing, and the evaluation does not separate their effects from differences in arm geometry. Counts include single-frame events without a dwell requirement. Prepared hardware replays establish no absolute task-matched torque benefit, temperature change or actuator-endurance improvement, and no result establishes attainment of an independently specified external target.

Future evaluation should measure external task positions in the robot's coordinate frame and identify contact events independently. This would connect the arm-motion findings to manipulation performance. Further work should check collisions during motion generation and compare reconstructed with reach-normalized targets on hardware using the same joint-angle method. Measuring time spent in saturation would distinguish brief events from sustained commands. Independent validation of the estimated arm geometry would also strengthen the interpretation.

\section{Conclusion}

A wrist target formed from demonstrated arm directions reflects the angle between those segments even without posture matching. With the joint-angle method fixed, reach normalization reduces saturated recordings from 8 to 6 on the reaching arm and from 14 to 5 on the supporting arm across 25 recordings, and eliminates saturation across the 20 repetitions of the forward-reach sweep. The bricklaying and lifting comparison is repeated with a second participant. Hardware replay shows within-run tracking and torque associations at saturated frames. Cross-body reach and attachment clearance identify limits of radial adjustment. These findings support evaluating how wrist positions are calculated alongside posture preferences and the requirements of the manipulation task.

\balance
\bibliographystyle{IEEEtran}
\bibliography{references}

\end{document}